\documentclass{sigchi}

\toappear{Permission to make digital or hard copies of all or part of this work for personal or classroom use is      granted without fee provided that copies are not made or distributed for profit or commercial advantage and that copies bear this notice and the full citation on the first page.} 

\usepackage{balance}  
\usepackage{graphics} 
\usepackage{txfonts}
\usepackage{times}    
\usepackage{url}      
\usepackage{color}
\usepackage{textcomp}
\usepackage{booktabs}
\usepackage{ccicons}
\usepackage{todonotes}
\usepackage{tabularx}
\usepackage{tabulary}
\usepackage{authblk}
\usepackage{enumitem}
\usepackage[normalem]{ulem}
\useunder{\uline}{\ul}{}
\usepackage[pdftex]{hyperref}
\PassOptionsToPackage{hyphens}{url}\usepackage{hyperref}
\makeatletter
\def\url@leostyle{%
 \@ifundefined{selectfont}{\def\UrlFont{\sf}}{\def\UrlFont{\small\bf\ttfamily}}}
\makeatother
\usepackage{cite}

\def\pprw{8.5in}
\def\pprh{11in}

\definecolor{linkColor}{RGB}{6,125,233}
\hypersetup{%
  pdftitle={SIGCHI Conference Proceedings Format},
  pdfauthor={LaTeX},
  pdfkeywords={SIGCHI, proceedings, archival format},
  bookmarksnumbered,
  pdfstartview={FitH},
  colorlinks,
  citecolor=black,
  filecolor=black,
  linkcolor=black,
  urlcolor=linkColor,
  breaklinks=true,
}

\begin{document}

\title{Evaluating semantic models with word-sentence relatedness}

\author[ ]{\rm Kimberly Glasgow}
\author[ ]{\rm Matthew Roos}
\author[ ]{\rm Amy Haufler}
\author[ ]{\rm Mark Chevillet}
\author[ ]{\rm Michael Wolmetz*}
\affil[ ]{The Johns Hopkins University Applied Physics Laboratory}
\affil[ ]{Laurel, MD}
\maketitle

    

\begin{abstract}
Semantic textual similarity (STS) systems are designed to encode and evaluate the semantic similarity between words, phrases, sentences, and documents. One method for assessing the quality or authenticity of semantic information encoded in these systems is by comparison with human judgments. A data set for evaluating semantic models was developed consisting of 775 English word-sentence pairs, each annotated for semantic relatedness by human raters engaged in a Maximum Difference Scaling (MDS) task, as well as a faster alternative task. As a sample application of this relatedness data, behavior-based relatedness was compared to the relatedness computed via four off-the-shelf STS models: n-gram, Latent Semantic Analysis (LSA), Word2Vec, and UMBC Ebiquity. Some STS models captured much of the variance in the human judgments collected, but they were not sensitive to the implicatures and entailments that were processed and considered by the participants. All text stimuli and judgment data have been made freely available. 
\end{abstract}

\subsection*{Corresponding Author}
Michael Wolmetz, \textit{michael.wolmetz@jhuapl.edu}

\section{Introduction}

There is some disagreement about what is meant by the terms \textit{conceptual knowledge} and \textit{semantic memory}, but most characterizations involve the encoding of beliefs or propositions about concepts (e.g. objects, actions, properties, etc.), the ability to organize those concepts into useful sets, and the relationships within and between sets \cite{patterson_where_2007,rogers_revisiting_2015,tyler2001towards}. From the perspective of the cognitive sciences, distinctions are typically made between the knowledge or memories closely tied to an individual's personal experiences, and those beliefs generally held in common by many individuals across a society, or many societies. Use of the term \textit{conceptual knowledge} is often biased toward these more universal beliefs and knowledge about concepts. 

Conceptual knowledge is studied by several related but distinct communities. In cognitive science, cognitive psychology, cognitive neuroscience, and computational linguistics, conceptual knowledge is typically studied in the context of discovering how concepts are learned, represented, and applied by humans. In natural language processing (NLP) and information retrieval, conceptual knowledge is often studied in the context of automated systems designed to make semantic inferences and retrieve information based on conceptual information.  The semantic models that drive these semantic textual similarity (STS) systems are learned from different properties or statistics extracted from text corpora (e.g. Wikipedia) or from the different relationships extracted from structured or taxonomic lexical knowledge databases (e.g. WordNet). 

Across all of these communities, ground-truths about concepts are necessary for testing theories and evaluating models or systems. These ground-truths can be challenging to generate because the attributes or dimensions of what should be considered part of a ground-truth may be dependent on the particular theory, model, or system being evaluated. One general solution to this problem has been to abstract away from specific attributes in favor of ground-truths about the relationships between concepts \cite{rips_semantic_1973}. Models can then be evaluated in terms of how well they predict either the semantic distances between concepts \cite{carlson2014emergence} or the structures or networks estimated from those distances\cite{steyvers2005large}. These distances or structures can be estimated from behavior-based measures like ratings or response times, corpus metrics, or relationships encoded in knowledge bases. 

A variety of semantic similarity and relatedness data sets have been used for these purposes. Many of the data sets used for NLP and evaluating STS systems are public, including those adopted by the International Workshop on Semantic Evaluation (Semeval, e.g. \cite{marelli_semeval-2014_2014}). Many of the data sets used in the cognitive sciences, though often reported on in published studies, have not been publicly released. 

Here we present a new data set for evaluating models of conceptual knowledge based on the relatedness between words and sentences. It consists of 25 target concepts (i.e. words) that span several semantic dimensions (objects, actions, settings, roles, states, and events), and parts of speech (nouns, verbs, and adjectives). Each of the 25 target words was paired with 31 simple sentences written and selected to elicit varying degrees of perceived relatedness to the target word, for a total of 775 word-sentence pairings. For example, the 31 sentences paired with the target word \textit{family} are shown in Table \ref{family_sentences}. 

\begin{table}[htbp]
\centering
\begin{tabular}{ll}
    \toprule
    \textbf{Sentence ID} & \textbf{Sentences for \textit{family}} \\
    \midrule
1  & The family was happy.                        \\
2  & The family played at the beach.              \\
3  & The family survived the powerful hurricane.  \\
4  & The wealthy family celebrated at the party.  \\
5  & The politician visited the family.           \\
6  & The parent watched the sick child.           \\
7  & The priest approached the lonely family.     \\
8  & The parent visited the school.               \\
9  & The parent shouted at the child.             \\
10 & The parent took the cellphone.               \\
11 & The couple planned the vacation.             \\
12 & The happy couple visited the embassy.        \\
13 & The parent bought the magazine.              \\
14 & The couple laughed at dinner.                \\
15 & The couple read on the beach.                \\
16 & The wealthy couple left the theater.         \\
17 & The happy child found the dime.              \\
18 & The child broke the glass in the restaurant. \\
19 & The child gave the flower to the artist.     \\
20 & The child held the soft feather.             \\
21 & The angry child threw the book.              \\
22 & The girl dropped the shiny dime.             \\
23 & The actor gave the football to the team.     \\
24 & The commander listened to the soldier.       \\
25 & The editor drank tea at dinner.              \\
26 & The soldier crossed the field.               \\
27 & The judge met the mayor.                     \\
28 & The beach was empty.                         \\
29 & The artist drew the river.                   \\
30 & The doctor stole the book.                   \\
31 & The window was dusty.                       \\
\bottomrule
\end{tabular}
\caption{The set 31 sentences paired with the target concept \textit{family}. Sentences are ranked from most related to the concept family to least related to the concept family, based on relatedness scores averaged across 47 participants.}
\label{family_sentences}
\end{table}

Some sentences are clearly related to the concept family, others appear somewhat related or slightly related, and others have no clear or obvious relationship to the target concept. These relationships were tested experimentally using a Maximum Difference Scaling procedure \cite{furlan_maximum_2014}, and replicated using a simpler procedure. As a sample application, the resulting ground-truth ratings were compared to the outputs of four commercial off-the-shelf (COTS) STS systems (n-gram, LSA, Google word2vec, and UMBC Ebiquity), and two baseline STS models to help interpret performance levels. 

This data set was originally generated for use with neural data as part of the Intelligence Advanced Research Projects Activity (IARPA) Knowledge Representation in Neural Systems (KRNS) Program.  While results are not reported here, the data set was used to evaluate whether semantic relatedness was encoded in the neural responses evoked by these sentences, as measured by functional Magnetic Resonance Imaging. All stimuli and results have been made available in the ancillary files published with this article.

\section{Methods \& Materials}

\subsection{Stimuli}
\textbf{Target concepts.} Table \ref{TCtable} lists the 25 target concepts included. Each target concept is restricted to a specific sense or meaning, adapted from WordNet senses for the term \cite{miller_wordnet:_1995}.  The set of target concepts consists of nouns, verbs and adjectives, and is biased away from abstract or uncommon concepts, and toward vivid, imageable \cite{paivio_concreteness_1968}, and concrete concepts. The mean concreteness for target concepts is 4.18 on the 5-point scale reported by Brysbaert \cite{brysbaert_concreteness_2014}. Target concepts were selected to span the six semantic dimensions described below. 

\begin{enumerate}
\item \textbf{Objects:} things that physically exist.  They may be animate or inanimate, natural or artifactual (man-made).  Objects will commonly have physical substance, or be detectable by human senses. These often take the form of nouns.  Objects may be count nouns or mass nouns.  Examples: rabbit, hammer, pear, clock, water.
\item \textbf{Actions:} things that are done or experienced (felt or sensed), typically by people or other living things.  These often take the form of verbs.  Actions may involve moving, perceiving, feeling, creating, and so on.  Examples: walks, hears, eats, builds, cooks.
\item \textbf{Settings:} where or when things happen.   They may be places, spaces, or times. Indoor or outdoor locations, seasons, and times of day are appropriate.  Examples:  dining room, plaza, winter, morning.
\item \textbf{Roles:} what people do or who they are.  They include vocations and professions, kinship, and social roles. Examples: athlete, victim, friend, brother, coach, plumber.
\item \textbf{States, properties, conditions, and emotions:} concepts that typically describe or characterize.  They include human emotions, physical properties, conditions, colors, and so on. They may take the form of adjectives. Examples: dry, red, damaged, sad.
\item \textbf{Events and activities:} things that happen (e.g. human organized or natural events) or activities that are engaged in (e.g. hobbies).  Examples: wedding, tornado, parade, tennis match.
\end{enumerate}

\begin{table*}[htb]
\centering
\begin{tabularx}{\textwidth}{l l l X}
\toprule
\textbf{Target} & \textbf{POS} & \textbf{Dimension} & \textbf{Sense}\\
                                                                                                                                               \textbf{Concept}
                                                                                                                                                    \\ \midrule
family                  & N            & Role               & primary social group; parents and children                                                                                                                           \\
school                  & N            & Setting            & a building where young people receive education                                                                                                                      \\
small                   & A            & State              & limited or below average in number or quantity or magnitude or extent                                                                                                \\
speak                   & V            & Action             & exchange thoughts; talk with                                                                                                                                         \\
break                   & V            & Action             & destroy the integrity of, usually by force; cause to separate into pieces or fragment                                                                                \\
trial                   & N            & Event              & legal proceedings consisting of the judicial examination of issues by a competent tribunal, including the determination of  innocence or guilt by due process of law \\
protest                 & N            & Event              & the act of protesting; a public (often organized) manifestation of dissent                                                                                           \\
lawyer                  & N            & Role               & a professional person authorized to practice law; conducts lawsuits or gives legal advice                                                                            \\
doctor                  & N            & Role               & a licensed medical practitioner                                                                                                                                      \\
walk                    & V            & Action             & use one’s  feet to advance, advance by steps                                                                                                                         \\
computer                & N            & Object             & a machine for performing calculations automatically                                                                                                                  \\
spring                  & N            & Setting            & the season of growth                                                                                                                                                 \\
tourist                 & N            & Role               & someone who travels for pleasure                                                                                                                                     \\
eat                     & V            & Action             & take in solid food                                                                                                                                                   \\
magazine                & N            & Object             & product consisting of a paperback periodic publication as a physical object                                                                                          \\
angry                   & A            & State              & feeling or showing anger                                                                                                                                             \\
park                    & N            & Setting            & a large area of land preserved in its natural state as public property, a piece of open land for recreational use                                                    \\
wealthy                 & A            & State              & having an abundant supply of money or possessions of value                                                                                                           \\
storm                   & N            & Event              & a violent weather condition with strong winds,  commonly featuring precipitation, thunder and lightning                                                              \\
soccer                  & N            & Event              & a football game in which two teams of 11 players try to kick or head a ball into the opponent goal                                                                   \\
kick                    & V            & Action             & strike, drive or propel with the foot                                                                                                                                \\
bird                    & N            & Object             & warm-blooded egg-laying vertebrates characterized by feathers and forelimbs modified as wings                                                                        \\
teacher                 & N            & Role               & a person whose occupation is teaching                                                                                                                                \\
dog                     & N            & Object             & a member of the genus Canis (probably descended from the common wolf) that has been domesticated by man since prehistoric times                                      \\
door                    & N            & Object             & a swinging or sliding barrier that will close the entrance to a room or building or vehicle                                                                          \\ \bottomrule
\end{tabularx}
\caption{The set of target concepts and associated metadata, including parts of speech (POS), noun (N), verb (V), or adjective (A); dimension; and sense. }
\label{TCtable}
\end{table*}

\subsection{Comparison sentences}
Each of the 25 target concepts was paired with 31 comparison sentences. Many comparison sentences were used across multiple target concepts, and in total, 240 distinct comparison sentences were generated (listed in the accompanying downloadable data). 

The specific set of 31 sentences associated and tested with a given target concept was constructed and selected to represent a range of semantic relatedness between the target concept and the selected comparison sentences. To achieve this range of relatedness, some sentences included the target concept, some sentences included at least one concept related to the target concept, and the remaining sentences did not contain any concepts thought to be highly related to the target concept. Where grammatically and semantically feasible, target concepts or related concepts appeared in the subject for some sentences and in the predicate for others, and occurred with or without adjectival modifiers (for nouns).

\section{Behavioral Procedure}
Human judgments were collected via a web-based application to assess the semantic relatedness between target concepts and comparison sentences. First, judgments were collected and analyzed using a Maximum Difference Scaling (MDS) or Best-Worst scaling paradigm \cite{louviere_best-worst_1991}. Because MDS sometimes requires very large numbers of responses, we also tested the applicability of a second paradigm, thought to be simpler and less resource-intensive ranking paradigm. 

\begin{figure}
\includegraphics[width=1.0\columnwidth]{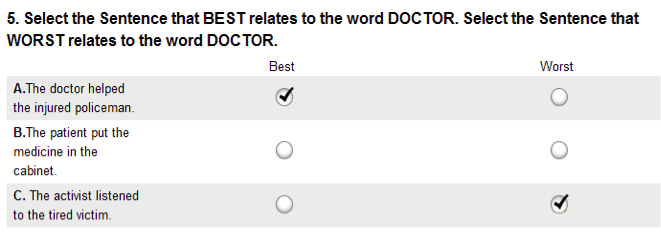}
\caption{One trial of the MDS paradigm for the target concept doctor.  The participant selects the sentence that best relates to the target concept, and the sentence that worst relates to the target concept. Each participant responded to 155 of these trials for the target concept doctor, and a total of 3875 of these trials in total.}
 ~\label{fig:figureMDSparadigm}
\end{figure}

MDS is a discrete choice technique used to evaluate relative importance or preference by asking participants to choose the “best” and “worst” options from a set of presented items.  A variant of MDS was used in which three sentences were presented with a target concept on each trial, as shown in Figure \ref{fig:figureMDSparadigm}. Participants were asked to consider the relatedness between the target concepts and the three sentences, and select the sentence that best relates to the target concept, and the sentence that worst relates to the target concept. While the terms similarity and relatedness are at times used interchangeably, and at other times used to refer to different relationships, the word “related” was used in all instructions given to participants, and so all results should be interpreted in that context. 

Responses from 155 of these trials were necessary to reconstruct relatedness scores for each of the comparison sentences for a single target concept, and a total of 3875 trials were collected for each participant. Presentation order was randomized across participants, and participants were permitted to log in and out of the application from the remote location of their choice, and complete the trials over as many sessions as necessary over the course of two weeks. 

The relatedness between a particular target word t and comparison sentence s was computed via counting analysis: the proportion of times each s was selected best for t (i.e. most related) less the proportion of times \textit{s} was selected worst for \textit{t} (i.e. least related):
 	
 	\begin{center}
 	$Relatedness_{s,t} = (s_{t,best}-s_{t,worst})/s_{t,total}$
 	\end{center}
 
In this way, similarity scores ranged between -1 (not similar) and 1 (very similar), and could be used to rank the sentences from the most related (highest score) to the least related (least related).  All further comparisons were done using the mean ranks of the sentences across participants.  

The very precise scores produced by MDS require large amounts of data collected over many trials. To assess whether this degree of precision was necessary, an alternative paradigm was explored in which participants were presented with all 31 sentences at once (in random order) and asked to rank them in terms of their relatedness to the target concept (from most related to least related) by dragging and dropping them into a list. Participants repeated this free-ranking procedure to produce 25 ranked lists of sentences: one for each of the 25 different target concepts.

\section{Participants}

Participants were recruited via flyers posted at local university campuses, and earned \$10 per hour for participation in accordance with a protocol approved by The Johns Hopkins Medical Institutions Institutional Review Board. All participants were fluent in English, had a high school diploma or equivalent, had normal or corrected to normal vision, and did not have any reading disorders and language impairments, as self-reported. Fifty-five participants (mean age = 24) completed the MDS task for all 25 concepts using a web-based application.  The data collected was analyzed for outliers, and eight of the 55 participants produced responses that deviated by more than three standard deviations from group means. These participants were removed from further analysis; 20 of the remaining 47 subjects also completed the free-ranking paradigm. 

\section{STS Systems}

\begin{figure*}[bth]
\centering
\includegraphics[width=0.9\textwidth]{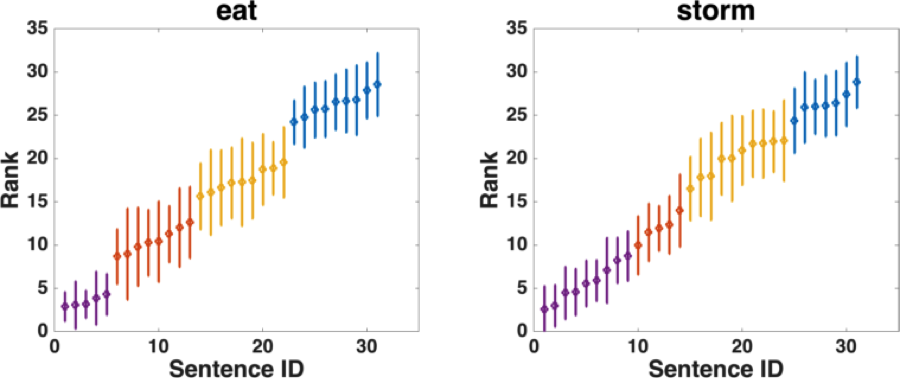}
\caption{Sentences for target concepts eat and storm clustered based on mean ranks using k-means clustering with k=4. Colors represent cluster assignments; error bars represent one standard deviation.}
 ~\label{fig:figureTcEatStorm}
\end{figure*}
 
The relatedness rankings produced by human participants were compared to relatedness rankings produced by four STS models, as well as two baseline text-based models to aid in interpretation. Each of the models listed below were run on the 25 target concept lists, and each produced 25 ranked lists of comparison sentences (just as the participants did). These ranked lists were compared to the behavior-based ranked lists using Spearman's rank correlation, and mean Spearman ρ values are reported. All analyses were run in Matlab.
\begin{enumerate}
\item \textbf{N-gram:} The basic n-gram model assumes that words that are related in meaning will occur in close proximity to one another \cite{brown_class-based_1992-1}. In practice, a matrix of co-occurrence frequencies between pairs of content words was generated, where co-occurrence frequency refers to the frequency with which two words appear within a five-word window of one another in the Corpus of Contemporary American English (COCA)\cite{davies_coca._2008-1}. In this way, co-occurrence frequency was used as a measure of semantic relatedness. 
More specifically, the matrix of co-occurrence frequencies was constructed as follows: A raw count matrix was generated in which each row and column entry contained a count of the number of times the two indexed words were both found within a five-word window.  The counts along the diagonal of the matrix were replaced by the total number of occurrences of the indexed word in the five-word windows – that is, each time a word was observed, it was counted as being paired with itself. The columns of the matrix were then normalized by their sums, resulting in a matrix of relative co-occurrence frequencies.  Similarity scores between target words and sentences were computed by summing the [target word, sentence word] elements of the matrix for all words in the sentence.
\item \textbf{LSA:} Latent Semantic Analysis is based on the principle that words that are related in meaning will occur in related texts or documents \cite{deerwester_indexing_1990-1}. Similarity was computed as the cosine distance between entries in the low-rank matrix. Analyses were run both with and without the use of a POS tagger, but POS information did not consistently change performance.
\item \textbf{Word2Vec:} The Word2Vec system was built by Google using a neural network and a skip-gram model\cite{mikolov_distributed_2013-1}. Skip-gram models are trained to predict the surrounding context given a target word. In this way, the similarity between two Word2Vec representations is computed as the cosine distance between vectors, and is associated with the similarity between the local contexts they could appear in based on the model.

\item \textbf{UMBC Ebiquity (Ebiquity):} The University of Maryland Baltimore County (UMBC) Ebiquity team provided its STS for inclusion in this comparison. The Ebiquity system fuses corpus-based (i.e. LSA) and knowledge-based (i.e. WordNet) methods to produce semantic similarity scores \cite{han_umbc_2013}. To augment the POS-tagged LSA-based semantic similarity, the WordNet database of semantic and lexical relations is used to produce an alternative estimate of similarity based on the distance between two words or concepts in the WordNet taxonomy (e.g. couple and family are similar because they both appear as instances of social groups within the WordNet hierarchy). The corpus-based and knowledge-based similarities are then weighted based on a training process to produce a single similarity score. In this way, the WordNet taxonomy effectively fills in certain gaps in distributional information. 

Ebiquity is specifically designed to compare words and sentences and compose a meaning for a sentence or phrase from the meanings of its constituent words.  The other models had not been optimized for this task, and so some method for comparing sentences to words had to be adopted. For the vector-based approaches reported here, we adopted similar methods to the LSA component of the Ebiquity system for comparing sentences to words: average the normalized individual word vectors from a comparison sentence, then compute the cosine distance between the target concept vector and the averaged sentence vector to produce a relatedness score between a target concept and a comparison sentence. This method of comparing words to sentences performed best amongst several alternatives tested.
\item \textbf{Word-spotting Baseline (WordSpot):} The word-spotting model adopted here as a baseline is an oracle that can only determine whether a comparison sentence contains the target concept or not. Based on this, the word-spotting model is constrained to rank target-containing sentences above sentences that do not include the target, but beyond this constraint, ranks are random. WordSpot was simulated 1000 times for each target concept, and correlations at the 95\% confidence interval were reported. 
\item \textbf{Relatedness-spotting Baseline (RelSpot):} This is similar to the word-spotting baseline model, but the relatedness-spotting model is an oracle that classifies words into one of three categories: targets, words related to the target, and unrelated words. Based on this, the relatedness-spotting model is constrained to rank target-containing sentences above sentences containing words related to the target, and these sentences, in turn, are constrained to be ranked above sentences comprised of unrelated words only. As with the WordSpot model, beyond this constraint, rankings are random.  Note that the ranking heuristic used by RelSpot is often consistent with the behavioral responses, but participants do rank comparison sentences with related words higher than some target-containing sentences at non-trivial rates. RelSpot was simulated 1000 times for each target concept, and correlations at the 95\% confidence interval were reported. 
\end{enumerate}

\section{Results and Discussion} 

Data collected from both the MDS task and free-ranking task were used to produce group rankings for the sentences associated with each target concept (see ancillary files). As expected, participants often ranked sentences that contained the target concept higher than sentences that did not, and often ranked sentences that contained a concept related to the target concept higher than sentences that did not contain the target concept or any related concepts, but these patterns were not absolute. For example (as shown in Table \ref{family_sentences}), the sentence \textit{The parent watched the sick child} was considered more related to the target concept family as compared to the sentence \textit{The priest approached the lonely family}, suggesting that participants considered the larger context when evaluating relationships between target concepts and comparison sentences. 
K-means clustering for k=3 to 5 was run on the mean rank data (as well as the mean ranks and variances) to assess whether relatedness varied continuously or discontinuously across sentences.  For some target concepts, the mean sentence ranks across participants were clearly clustered by relatedness to the target concept, while for other target concepts, the sentence rankings varied more continuously (as illustrated in Figure \ref{fig:figureTcEatStorm}).  For example, data collected for the target concept \textit{eat} suggests that comparison sentences can be grouped into four somewhat separable classes: sentences highly related, related, somewhat related, and unrelated to the target concept. Alternatively, data collected for the target concept \textit{storm} do not clearly fall in to these relatedness classes.

Cross-test reliability was calculated for the subset of participants who completed both the MDS task and the free-ranking task (N=20). While the MDS task produces scores ranging from -1 to 1 associated with relatedness, the free-ranking tasks only produce ranks, and so only ranks were compared between tasks. The mean rank difference between MDS and free-ranking group rankings was 1.3: on average, a comparison sentence’s rank (of 31) on the MDS task was 1.3 positions away from that sentences rank according to the free-ranking task results, which is similar to the within-task reliability observed for these tasks. Given that the free-ranking task was on the order of seven times faster to complete, it may be a preferred method for collection of relatedness judgments.

\section{STS System Comparison}

The sentence rankings from each of the STS models were correlated with the mean MDS rankings from participants, and the results are reported in Table \ref{syscompar}. All models except for the n-gram model unambiguously outperformed the word-spotting baseline, but only the Ebiquity and Word2Vec systems outperformed the relationship-spotting benchmark system. These results suggest that all models except the n-gram model are capable of conceptual inference beyond simple word-spotting. The sub-baseline performance of the n-gram model might marginally improve with a larger corpus, but the interpretation would not change: higher-order models which can encode more abstract contextual information are consistently more predictive of human judgments as compared to the lower-order models that encode only surface information about local context. Indeed, both the Word2Vec and Ebiquity models consistently ranked sentences containing words related to the target above sentences that did not contain any related words. Given their beyond-benchmark performance, these models also appear capable of inferring some additional and useful conceptual structure.  

\begin{figure}[htb]
\centering
\includegraphics[width=\columnwidth]{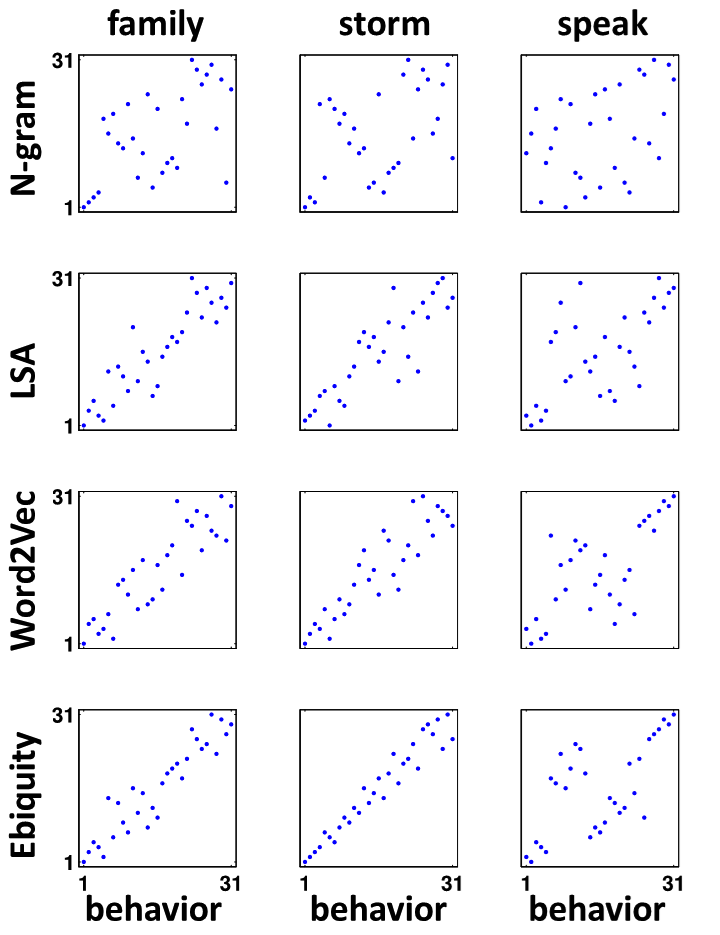}
\caption{Model comparisons with behavior for rankings of three representative target concepts. For the target concept \textit{family}, the n-gram model performs substantially worse than each of the other models. For \textit{storm}, the Ebiquity system performs substantially better than each of the other models. For \textit{speak}, all models perform similarly poorly at predicting behavioral responses. The collective poor performance for \textit{speak} was partially due to the failure across models to infer that to \textit{speak} is an important component of \textit{interview} and \textit{negotiate}, which are content words appearing in many of the comparison sentences associated with \textit{speak}.}
 ~\label{fig:figureTcbehaviorthumbs}
\end{figure}

Performance was not uniform across all target concepts. For some target concepts, STS results were quite similar to participant ranking (e.g. \textit{storm}), and for others, there were large discrepancies (e.g. \textit{speak}). Results for three representative target concepts compared to aggregate behavioral rankings are shown in Figure \ref{fig:figureTcbehaviorthumbs}.  To better understand the relationship between human and system ratings, the Ebiquity results were further examined at the individual sentence level.

\begin{table}[]
\centering

\begin{tabularx}{\linewidth}{l cX}
\toprule
\textbf{System}   & \textbf{Correlation with Behavior (\(\rho\))}   \\ \midrule
N-gram   & 0.39                          \\
WordSpot & 0.46                          \\
LSA      & 0.72                          \\
RelSpot  & 0.76                          \\
Word2Vec & 0.78                          \\
Ebiquity & 0.80                          \\ \bottomrule
\end{tabularx}
\caption{Correlations with judgment-based rankings across all Target Concepts. The WordSpot and RelSpon baseline models are shown in gray; correlations for these baseline models reflect the 95\% Confidence Interval of performance based on a 1000 iteration Monte Carlo simulation. }
\label{syscompar}

\end{table}

For many comparison sentences and for most target concepts, the Ebiquity model produced rankings quite similar to those produced by the human raters, but there were notable exceptions.  For several target concepts, some sentences were ranked more than 15 positions (of 31) higher or lower by the STS system than by human judgment.  Some of these discrepancies suggest incompatibilities between the human task (i.e. relatedness judgments) and the STS task (semantic similarity). Other discrepancies suggest true limitations of the STS systems.

A number of these large discrepancies occurred when concepts were related by contrasts in meaning (i.e. antonyms). For example, some of the largest discrepancies between the human and Ebiquity rankings involved the target concept \textit{small}.  \textit{Small} is a gradable antonym, on the opposite end of a pole with \textit{big}.  Because \textit{big} and \textit{small} are both adjectives describing two poles of size, they might appear in similar contexts in corpora.  As a result, the sentence \textit{The big horse drank from the lake} was ranked the seventh most similar sentence to the target concept small (i.e. 7/31) by the Ebiquity system, and 30/31 as ``best related" to the target concept small by human participants. 

Multiple incompatible antonyms \cite{gairns_working_1993}, which are mutually exclusive terms that belong to a set (e.g., the seasons of the year: spring, summer, fall, winter) showed similar patterns of results.  This was observed with \textit{spring} (a target concept) and other seasons that appear in comparison sentences like \textit{winter} (i.e. \textit{The park was empty in winter}: Human rank 25/31; Ebiquity rank 7/31).  These and other similar results highlight the different conceptions of similarity and relatedness, as well as the limitations inherent in low-dimensional similarity and relatedness scores. Different instructions or a different task may have induced behavioral results more in line with the Ebiquity rankings, but this is also a relatively unexplored distinction with respect to STS research.

Other discrepancies between the human and STS results involved relationships that the STS model was not sensitive to, and that would not be remedied by alternative instructions or tasks. In these cases, the comparison sentences do not include the target concept, but instead, include concepts that prototypically embody the target concept or evoke it through entailment or implicature. 

Entailments are truths that logically must also hold, given that an initial statement was true, as in ``Bob was murdered" entails ``Bob is dead."  Similarly, implicatures are highly likely to be true, but not definitively so \cite{grundy_doing_2013}. They convey meaning indirectly and are understood implicitly. The statement ``Bob was buried" suggests ``Bob is dead" via implicature, yet it is possible, though extremely unlikely, for someone to be buried alive. To see how the Ebiquity model is insensitive to important entailments, consider the target concept small and the human and Ebiquity rankings of comparison sentences that reference prototypically small objects, as shown in Table \ref{small}.


Participants clearly associated dimes and the concepts of finding or losing dimes with small, but the STS model did not. Participants rated sentences about dimes as more related to small than sentences containing the word small, but in reference to items that are not protypically small. The STS did not mimic this behavior, nor was it sensitive to the smallness of mice.

\begin{table}[hbtp]
\centering
\begin{tabularx}{\linewidth}{l cX cX}
\toprule
         & \multicolumn{2}{c}{\textbf{Rank (of 31)}} \\ 
         \cmidrule(r){2-3}
        
{\small\textbf{Comparison sentence (\textit{small}) }}
                                   & \textit{Human}         & \textit{Ebiquity}         \\
The girl dropped the shiny dime.   & 5             & 30               \\
The dime was new.                  & 7             & 23               \\
The mouse ran into the forest.     & 11            & 22               \\ \bottomrule
\end{tabularx}
\caption{A subset of human relatedness and Ebiquity similarity rankings with respect to the target concept \textit{small}.}
\label{small}
\end{table}

Results from other target concepts similarly demonstrated the lack of STS model sensitivity to entailment and implicature. For the target concept speak, participants considered sentences with concepts like negotiate and interview: acts which involve speaking. Participants found these sentences highly related to speak, but the STS model did not: \textit{The commander negotiated with the council} (Human rank 7/31; Ebiquity rank 24/31) and \textit{The author interviewed the scientist after the flood} (Human rank 6/31; Ebiquity rank 18/31). For the target concept door, participants recognize that leaving a building such as a theater implicitly requires passing through a door, while the Ebiquity system does not, as in \textit{The wealthy couple left the theater }(Human rank 10/31; Ebiquity rank 24/31).

Finally, there were more complex implicatures that participants incorporated into their judgments that the Ebiquity system did not. As an example, the target concept tourist was evaluated with the comparison sentence The minister visited the prison (Human rank 23/31; Ebiquity rank 8/31). While visiting a location is the essence of tourism, prisons are perhaps the antithesis of a vacation destination, and so participants modulated their interpretation accordingly. A more complete and human-like conceptual knowledge representation and interpretation process that is better able to incorporate entailment, implicature, and antonymy should be expected to perform better in these contexts.   Indeed, some automated approaches to text summarization and question-answering have considered these factors\cite{lloret_text_2012}.
Note that results from both the MDS and free-rank task were assembled into a collection of ancillary text files and have been made available for download. The files include lists of all concepts and sentences used in the tasks, along with word-sentence pairings for the 25 tested concepts. Mean ranks and standard deviations are reported for the two tasks individually, with sentences ordered by mean rank for each concept.

\section{Conclusion}

We have presented a data set for evaluating semantic models based on human behavioral rankings. 775 English word-sentence pairs were constructed to embed concepts in meaningful contexts and generate a range of word-sentence relationships. Word-sentence pairs were annotated for semantic relatedness by human raters, and ratings were found reliable across rating tasks. 

To illustrate the potential utility of this data set, results of the rating task were compared to several semantic textual similarity systems. Higher-order text systems like Word2Vec and UMBC Ebiquity often closely predicted human ratings, but failed to match human judgments in some cases. The first major disagreement between human and STS ratings arose from subtle differences between semantic relatedness and semantic similarity, and more attention may need to be paid to this distinction when developing evaluations and applications for semantic models.  The second major disagreement came from a failure of the semantic models to capture entailment, implicature, and context. Human relatedness judgments are shaped by understanding of context and the ability to infer entailed and implied information. Semantic models will need to be enriched with additional sources to more closely parallel human semantic judgments.

\section{Acknowledgments}

This research is based upon work supported by the Office of the Director of National Intelligence (ODNI), Intelligence Advanced Research Projects Activity (IARPA), via IARPA Contract No. 2012 – 12050800010. The views and conclusions contained herein are those of the authors and should not be interpreted as necessarily representing the official policies or endorsements, either expressed or implied, of the ODNI, IARPA, or the U.S. Government. The U.S. Government is authorized to reproduce and distribute reprints for Governmental purposes notwithstanding any copyright annotation thereon.

The authors also gratefully acknowledge the generous contribution of Dr. Lushan Han, Samsung Research America, and Dr. Tim Finin, University of Maryland, Baltimore County, in making the code for the Ebiquity STS system available, Mary Luongo in helping to collect behavioral responses, and Dr. Christine Piatko in discussing relevant research topics.

%
%
%
%
%
\balance{}

\bibliographystyle{SIGCHI-Reference-Format}
\bibliography{NIDBStimPaper}
\end{document}